\documentclass{article}

\PassOptionsToPackage{numbers, compress}{natbib}
\usepackage[final]{nips_2018}
\usepackage{color}

\usepackage[utf8]{inputenc} 
\usepackage[T1]{fontenc}    
\usepackage{hyperref}       
\usepackage{url}            
\usepackage{booktabs}       
\usepackage{amsfonts}       
\usepackage{nicefrac}       
\usepackage{microtype}      
\usepackage{graphicx}

\title{TIFTI: A Framework for Extracting Drug Intervals from Longitudinal Clinic Notes}

\author{
  Monica Agrawal   \textsuperscript{1, 2},
  Griffin Adams   \textsuperscript{1},
  Nathan Nussbaum \textsuperscript{1, 3},
  Benjamin Birnbaum   \textsuperscript{1}\\
  \textsuperscript{1} Flatiron Health, New York, NY \\
  \textsuperscript{2}   Massachusetts Institute of Technology, Cambridge, MA\\
  \textsuperscript{3}   New York University School of Medicine \\
}

\begin{document}

\newcommand{\prelabel}{\texttt{PRE}}
\newcommand{\midlabel}{\texttt{MID}}
\newcommand{\postlabel}{\texttt{POST}}
\newcommand{\algname}{TIFTI}

\maketitle

\begin{abstract}
Oral drugs are becoming increasingly common in oncology care. In contrast to intravenous chemotherapy, which is administered in the clinic and carefully tracked via structured electronic health records (EHRs), oral drug treatment is self-administered and therefore not tracked as well. Often, the details of oral cancer treatment occur only in unstructured clinic notes. Extracting this information is critical to understanding a patient’s treatment history. Yet, this is a challenging task because treatment intervals must be inferred longitudinally from both explicit mentions in the text as well as from document timestamps. In this work, we present \algname\ (Temporally Integrated Framework for Treatment Intervals), a robust framework for extracting oral drug treatment intervals from a patient’s unstructured notes. \algname\ leverages distinct sources of temporal information by breaking the problem down into two separate subtasks: document-level sequence labeling and date extraction. On a labeled dataset of metastatic renal-cell carcinoma (RCC) patients, it exactly matched the labeled start date in 46\% of the examples (86\% of the examples within 30 days), and it exactly matched the labeled end date in 52\% of the examples (78\% of the examples within 30 days). Without retraining, the model achieved a similar level of performance on a labeled dataset of advanced non-small-cell lung cancer (NSCLC) patients.
\end{abstract}

\section{Introduction}
The widespread adoption of electronic health records (EHRs) has raised hopes that data from routine clinical practice can be used to better understand the effectiveness of drugs in real-world settings. This is particularly important in cancer care, where treatment landscapes for many diseases have evolved rapidly~\cite{harrison2013real}. In renal cell carcinoma (RCC), for example, ten new targeted drugs have been approved since 2006, leaving clinicians in need of answers for how best to use these different treatments to care for their patients~\cite{bamias2017current}. Many of the new treatments for RCC and for other cancer types are oral therapies, which patients take at home. Unfortunately, exposure to oral therapies is generally not well captured in EHR structured data, but rather the information is buried in free text such as clinic visit notes~\cite{turchin2009comparison,wang2017clinical}. Manual abstraction of oral therapy exposure data by a human expert is possible but inefficient, so there is a need for automated approaches for extracting drug treatment information~\cite{matt2013retrospective}.

Most existing work on extracting drugs from EHRs has focused on discharge summaries, which represent a single point in a patient’s clinical course \cite{n2c2,sun2013evaluating,uzuner2010extracting,wang2017clinical}. In contrast, understanding the chronology of a disease such as cancer requires treatment information that is scattered longitudinally across clinic notes. The modeling task we address is: given the name of a drug and a sequence of time-stamped clinic visit notes, predict whether the patient took the drug and if so, generate the time interval over which the patient took it. Our contribution is a framework called \algname\ (Temporally Integrated Framework for Treatment Intervals), which approaches this problem by combining the output from two sub-tasks: a document-level sequence labeling task and a date extraction task. 

\section{Dataset} 
We conducted our study on the clinic visit notes of a set of 4,216 patients with metastatic RCC, pulled from the Flatiron Health  database, a longitudinal, demographically, and geographically diverse database derived from EHR data. Institutional Review Board approval was obtained prior to study conduct and informed patient consent was waived. Oral drug regimens, along with their start and end dates, were abstracted by experts via chart review \cite{berger2016opportunities}.\footnote{See Appendix for which drugs were included.} For the purpose of developing and validating an ML model, we used these labels as ground truth. The units of observation for our task were patient-drug pairs. Only pairs in which the clinic notes contained at least one mention of the drug (either by the generic or brand name) were considered. There were 8,259 such patient-drug examples from 172 different practices. Of these, the drug was actually taken in 4,410 (53\%) examples; in the rest, the drug was mentioned in the notes but not taken.

\section{Description of \algname}
Start and stop dates for oral drug regimens can be inferred from clinic notes in two ways. The first is via an explicit mention of an absolute or relative date, such as ``Patient started DRUG on 12/8/18'' or ``Patient discontinued DRUG two weeks ago.'' The second is via an implicit mention of whether the drug is currently being taken, such as ``Patient tolerating DRUG well'' or ``Patient no longer on DRUG.'' Both types of mentions are prevalent. In our data, we found that only about half of the start and stop dates could be found via explicit mentions, meaning that the rest must be inferred by anchoring implicit mentions to document timestamps. On the other hand, only 55\% of the start dates and 77\% of the end dates corresponded to a document timestamp, meaning that implicit mentions alone are also insufficient. Our analysis further found that different practices have significantly different ways of recording oral drug regimens.
This motivates the design of the \algname\ framework, which combines document-level sequence labeling (for implicit mentions) with free text date extraction and labeling (for explicit mentions).

\begin{figure}
  \centering
\includegraphics[width=\textwidth]{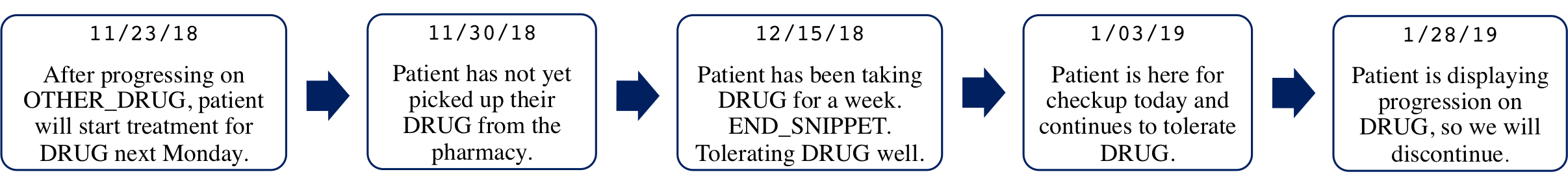}
  \caption{Toy example of processed document timeline. The start date should be  12/08/18 because the patient has been on the drug “for a week” as of the 12/15/18 note. The end date should be 1/28/19.}\label{timeline}
\end{figure}

\paragraph{Data Preprocessing} \algname\ starts with a series of preprocessing steps for each patient-drug example. 
First, it tokenizes the documents into sentences~\cite{bird2004nltk} and removes all documents and sentences without a mention of the drug.  Next, to remove the redundancy caused by ``copy-forwarding'' \cite{cohen2013redundancy}, it removes any sentence that occurred previously. Finally, to reduce feature sparsity and ensure that model features generalize across drugs, it replaces each mention of the drug (by any of its generic or brand name synonyms) with the placeholder DRUG and each mention of other commonly taken drugs with the placeholder OTHER-DRUG.  We refer to the list of condensed documents that are output by this preprocessing as the \emph{document timeline}. A toy example is provided in Figure~\ref{timeline}. 

\paragraph{Document Timeline Sequence Labeling}
The first task modeled by \algname\ is a sequence labeling task on the document timeline. Each document is labeled with one of three labels, depending on how the document timestamp relates to the ground truth interval: \prelabel\ if it is before the start date, \midlabel\ if it falls within the interval, or \postlabel\ if it is at or after the end date. A classification algorithm on this task will output for each document a probability distribution across these labels. To translate these probabilities to an interval prediction, the following algorithm can be used. First, the most probable sequence of labels is found such that no \midlabel\ precedes a \prelabel\ and no \postlabel\ precedes a \midlabel. Then, if there are no \midlabel\ or \postlabel\ documents, the algorithm outputs that the drug was not taken. Otherwise, the start date is set to the timestamp of the first document with a label of \midlabel\ and the end date to the timestamp of the first document with a label of \postlabel\ (assuming one exists). If we use the document timeline in Figure~\ref{timeline} to generate the labeled data for our training step, the first two documents would be labeled as \prelabel, the next two as \midlabel, and the final document as \postlabel. Assuming the correct document label sequence was predicted, the predicted start date would be 12/15/18 (timestamp of the first \midlabel\ document) and the predicted end date would be 1/28/19 (timestamp of the first \postlabel\ document). 

\begin{figure}
  \centering
\includegraphics[width=\textwidth]{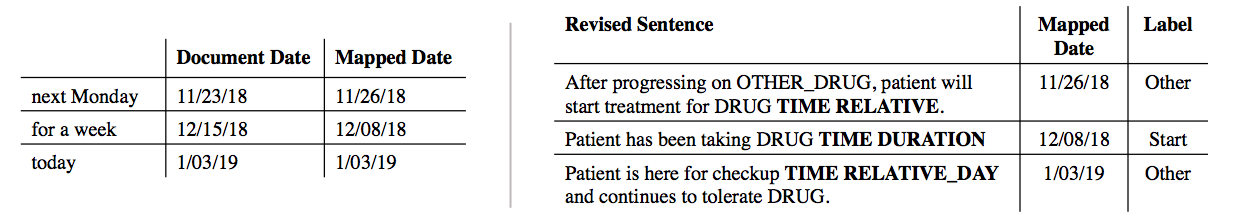}
  \caption{Time expression extraction, mapping, and labeling on the example from Figure~\ref{timeline}.}\label{extraction}
\end{figure}

\paragraph{Time Expression Classification}
The second task modeled by \algname\ is a joint date extraction and labeling task.  The first step of this process is conducted using a regular expression-based temporal tagger that categorizes possible time expression types into one of a few buckets, such as explicit dates (e.g. ``November 27'') and relative dates (e.g. ``last Tuesday'') and then calculates the date the expression is referring to, which we refer to as the \emph{mapped date}. The left side of Figure~\ref{extraction} shows this step on our toy example. The second step of the process is a three-class classification problem on the sentence containing the time expression to determine whether the mapped time refers to the beginning of a drug regimen, the end of a drug regimen, or neither. In each sentence, the time expression is replaced with TIME BUCKET-NAME. We refer to this new sentence as a \emph{revised sentence}. The right side of Figure~\ref{extraction} shows this step on our toy example. Since there are no explicit ground truth labels for whether the content of a revised sentence actually refers to the beginning or end of the regimen, labels for training are generated using a noisy proxy: whether the mapped date is within a small number of days (chosen as a hyperparameter and optimized on the development set) of the labeled start or end date. At evaluation time, for each time expression, the time expression classifier provides a probability estimate that the expression refers to a start date, end date, or neither.

\begin{figure}
  \centering
\includegraphics[width=\textwidth]{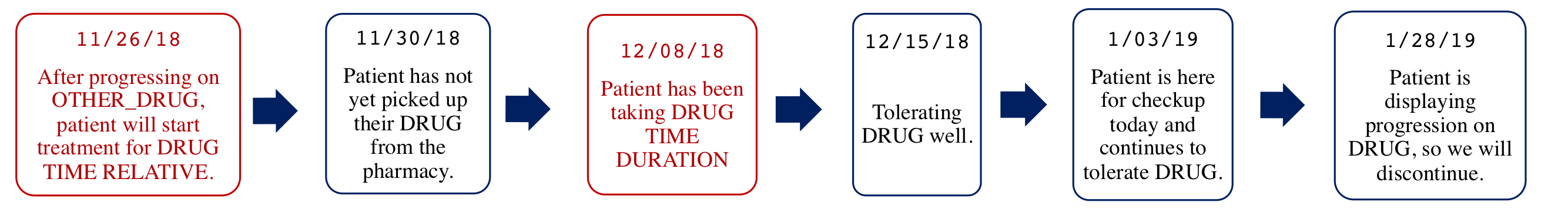}
  \caption{Toy example of simulated version of the document timeline from Figure~\ref{timeline}. Two pseudo-documents, indicated in red, were added for sentences that corresponded to time expressions.}\label{simulated}
\end{figure}

\paragraph{Integrated Prediction Task} \algname\ combines the output of the two sub-tasks. First, the time expression classification algorithm is run. If the maximum probability of an expression being either a start or an end date is above a threshold (chosen as a hyperparameter), \algname\ uses that start or end date for its predicted interval. Otherwise, it creates a new \emph{simulated document timeline}, where \emph{pseudo-documents} are added to the original document timeline. The contents of these pseudo-documents are the revised sentences from the time expression classification approach, and their timestamps are the mapped dates. A simulated document timeline for our toy example is shown in Figure~\ref{simulated}. \algname\ then runs an algorithm for the sequence labeling task on this simulated document timeline.

\section{Experimental Setup}
We used 80\% of our RCC examples for development and reserved 20\% for testing. The dataset was split such that there was no patient-level overlap between development and test sets. 

We measured the performance of the binary task of predicting whether the patient took the drug using the $F_1$ score. 
For true positive examples (those for which the model correctly predicted that the patient took the drug), we measured the agreement of start and stop dates as follows. Let $\mathrm{Start}_i(t)$ and $\mathrm{Stop}_i(t)$ be an indicator variable denoting whether for the $i^{\mathrm{th}}$ example, the predicted start or stop date matches the ground truth date within a window of $t$ days. For example, $\mathrm{Stop}_i(7)=1$ iff either the patient is still taking the drug and the model correctly identifies this, or the last-taken date identified by the model is within a week of the ground truth. To measure overall date agreement, we used $\mathrm{Start}(t)$ and $\mathrm{Stop}(t)$, defined to be the mean over the true positives of the $\mathrm{Start}_i(t)$ and $\mathrm{Stop}_i(t)$ values.


To remain flexible and sensitive to dataset size, the \algname\ framework does not specify the classification algorithm for either sub-task. We tried multiple algorithms for each. On the document timeline sequence labeling task, we saw the best performance with a bidirectional LSTM \cite{graves2005framewise} over documents featurized by ngrams. On the time expression classification task, we saw the best performance with a simple $\ell_2$-regularized logistic regression, also featurized by ngrams. These optimizations, along with other hyperparameter tuning, were performed using 5-fold cross validation over the development set, optimizing on a combination of the $F_1$ score, $\mathrm{Start}(0)$, and $\mathrm{Stop}(0)$. 

To perform well for rare drugs and generalize across diseases, \algname\ abstracts away the drug name during feature generation and models each drug independently. To test whether this design had the intended effect, we created a dataset of 8,388 advanced non-small cell lung cancer (NSCLC) examples (6,810 in the development set and 1,578 in the test set), using the same data preprocessing and feature generation process as for RCC. We then measured the performance on the NSCLC test set of the final \algname\ model trained on RCC and of a \algname\ model trained on NSCLC examples.

\section{Results}

\begin{table}
  \caption{Ablation study of \algname\ framework, applied to test set of 1,615 examples.}\label{results}
  \label{sample-table}
  \centering
  \begin{tabular}{llllll}
    \toprule
    Method     & $F_1$ Score   & Start(0) &  Stop(0) & Start(30) & Stop(30)\\
    \midrule
    Timeline Labeling & 0.943  & 23.8\%  & 51.4\%  & 82.5\% & \textbf{77.8\%} \\
    \hline 
    Simulated Timeline Labeling    & 0.943 & 41.0\%    & 52.2\% & 85.7\% & 77.2\% \\
        \hline 
    Expression Classification + \\ Timeline Labeling & \textbf{0.946}     & 44.7\%  & \textbf{52.4\%} & 83.6\%& 77.7\% \\
        \hline 
    Expression Classification + \\  Simulated Timeline Labeling (\algname)    & 0.944 & \textbf{45.8\%} & \textbf{52.4\%} & \textbf{85.9\%} & 77.6\% \\
    \bottomrule
  \end{tabular}
\end{table}

On the RCC test set, the full TIFTI model achieved an $F_1$ score of 0.944, a $\mathrm{Start}(0)$ score of 45.8\%, a $\mathrm{Stop}(0)$ score of 52.4\%, a $\mathrm{Start}(30)$ score of 85.9\%, and a $\mathrm{Stop}(30)$ score of 77.6\%.
We performed an ablation study (Table~\ref{results}) on our framework to demonstrate the value added by: (i) the simulated timelines and (ii) the explicitly cascaded expression classification. The explicitly cascaded models performed best, and the models with the simulated document timelines outperformed their counterparts with the original document timeline, both at 0 and 30 days, confirming that  pseudo-documents add useful context. This effect is only visible for the start date statistics, which is consistent with the fact that start dates were more likely than stop dates to be explicitly mentioned in text.

On the NSCLC test set, the model trained on the RCC data achieved an $F_1$ score of 0.936, a $\mathrm{Start(0)}$ score of 49.1\%, and a $\mathrm{Stop(0)}$ score of 57.1\%. This performance is comparable to the performance on the RCC test set and almost as high as the model trained on the NSCLC examples ($F_1$: 0.947, $\mathrm{Start}(0)$: 50.3\%, $\mathrm{Stop}(0)$: 57.8\%), indicating that the framework generalized as intended.

\section{Conclusion}

We introduced \algname, a framework for extracting the spans of drug regimens from longitudinal clinic visit notes. \algname\ predicts the treatment interval over a simulated patient timeline formed by combining the temporal information from both free text and document timestamps. In our experiments, it predicted approximately 80\% of dates within 30 days and generalized to a new type of cancer. This work is significant in that this framework could be applied to build EHR datasets that capture exposure to anti-cancer therapies. Such datasets can then be de-identified and analyzed to understand the real-world patient experience and develop insights that can impact cancer care.

\bibliographystyle{plainnat}

\section{Appendix: Included Oral Drugs}

For RCC, the oral drugs included were axitinib, cabozantinib, erlotinib, everolimus, lenvatinib, pazopanib, sorafenib, and sunitinib.
For NSCLC, the oral drugs included were afatinib, alectinib, brigatinib, ceritinib, crizotinib, erlotinib, gefitinib, osimeritinib, cabozantinib, dabrafenib, trametinib, vandetanib, and vemurafenib.

\end{document}